\documentclass[letterpaper]{article}
\usepackage{flairs}
\usepackage{times}
\usepackage{helvet}
\usepackage{courier}
\usepackage{graphicx}
\begin{document}
%
\title{Influencing Reinforcement Learning through Natural Language Guidance}
\author{Tasmia Tasrin \\
University of Kentucky\\
{\tt\small tta245@uky.edu}
\And 
Md Sultan Al Nahian\\
University of Kentucky\\
{\tt\small mna245@uky.edu}
\And
Habarakadage Perera\\
University of Kentucky\\
{\tt\small hmpe228@uky.edu}
\And
Brent Harrison\\
University of Kentucky\\
{\tt\small bha286@g.uky.edu}

 }
\maketitle
\begin{abstract}
\begin{quote}
Interactive reinforcement learning (IRL) agents use human feedback or instruction to help them learn in complex environments. 
Often, this feedback comes in the form of a discrete signal that’s either positive or negative. 
While informative, this information can be difficult to generalize on its own.
In this work, we explore how natural language advice can be used to provide a richer feedback signal to a reinforcement learning agent by extending policy shaping, a well-known IRL technique. 
Usually policy shaping employs a human feedback policy to help an agent to learn more about how to achieve its goal. 
In our case, we replace this human feedback policy with policy generated based on natural language advice. 
We aim to inspect if the generated natural language reasoning provides support to a deep RL agent to decide its actions successfully in any given environment. 
So, we design our model with three networks: first one is the experience driven, next is the advice generator and third one is the advice driven. 
While the experience driven RL agent chooses its actions being influenced by the environmental reward, the advice driven neural network with generated feedback by the advice generator for any new state selects its actions to assist the RL agent to better policy shaping.
\end{quote}
\end{abstract}

\section{Introduction}

Reinforcement learning (RL) is a machine learning approach that teaches agents to exhibit behaviors that maximize a numeric reward signal through trial-and-error.  
RL has proven that it can train agents in complex environments with unknown information. 
There are situations, however, where RL agents struggle to learn. 
For example, it is well known that environments with sparse reward signals can prove difficult for classic RL agents. 
In these situations, some researchers have sought to augment classic RL approaches with additional human knowledge in the way of direct feedback or instruction. 
These approaches, called interactive reinforcement learning (IRL) techniques, utilize this human knowledge to better enable agents to learn in especially complex environments. 
Typically, humans provide this knowledge by either providing demonstrations of positive behavior or by providing numeric feedback on the quality of the actions taken by the agent during training. 
These, however, can be difficult for humans to provide in a way that is useful to the RL agent, especially when provided by teachers with little or no machine learning or artificial intelligence expertise. 

To address this limitation of IRL, we explore the possibility of using natural human advice as a means to provide feedback to an IRL system.
Specifically, our approach involves using this advice to train a computational \textit{advice generator} which the agent can then use to determine the quality of potential future actions. 
This will enable the agent to receive targeted feedback commenting on the quality of actions or suggestions for future actions to take while allowing humans to provide feedback in a natural way. 
Our work extends previous work performed by Harrison \textit{et al.}~\cite{harrison2018guiding} in which they show that natural language can be used to help simple agents generalize knowledge to unseen environments. 
However, their approach utilizes highly structured language that was generated using a synthetic grammar, and they limited their study to grid world environments. 
In this work, we present a system for Automated Advice Aided Policy Shaping, or A3PS, an end-to-end system for training agents in complex environments that combines deep reinforcement learning techniques with generated advice trained on human advice given in natural language. 
This is done by modifying the \textit{policy shaping} algorithm, an IRL algorithm that learns from human critique~\cite{NIPS2013_5187}.

To evaluate this system, we explore its effectiveness using the arcade game Frogger.
In contrast to the work by Harrison \textit{et al.}, however, we evaluate our approach using a Frogger environment that utilizes a pixel state environment rather than a simpler grid-based one. 
We compare our approach against state-of-the-art RL baselines and show that the inclusion of natural language can significantly enhance learning even when rewards are sparse.

\section{Related Works}
The goal of IRL is to use human knowledge to help an autonomous agent learn in uncertain environments. 
One way that this can be done is by having a human teacher directly specify the reward function for an agent in various ways~\cite{reward1,reward2,reward3,tamer,humanteachers}. 
While these methods have proven effective, specifying a reward function directly can be difficult for humans as often it is unclear how reward signals directly translate into behaviors. 



To alleviate this, researchers developed methods for using human feedback to augment environmental reward. 
These approaches would use machine learning or deep learning methods to merge various forms of human feedback with environmental reward in a way that often balances between the two~\cite{Knox2010,knox12,dqn}. 
Ultimately, however, the goal of these approaches is to use human feedback to help an agent learn to maximize environmental reward. 

These approaches still have limitations as they do not take into account the fact that human feedback signals are often inconsistent. 
One explanation for this is that humans have their own policy for providing said feedback.
One way to address this limitation is to use human demonstrations of positive behavior to train an autonomous agent~\cite{demo1,demo2,demo3,demo4}. 
This enables the agent to see examples of desirable behavior and learn from them. 
This can be problematic for humans as it can be difficult to specify what types of demonstrations will best help an agent learn. 

Another option is to attempt to model how a human teacher provides feedback. 
This is the basis of the policy shaping algorithm~\cite{griffith2013policy,policy2}, which seeks to model the feedback policy of a human trainer and then combine it with a policy derived from an agent's experience in an effort to guide exploration. 
The ultimate goal is still to train an agent that maximizes environmental reward, but this better enables it to understand human feedback. 
While this method has proven effective in practice, it is limited to working with discrete feedback. 
In this work, we aim to extend it to better incorporate natural language instructions. 

As mentioned previously, our A3PS algorithm was inspired by the work of Harrison~\textit{et al.}~\cite{harrison2018guiding} in which they show that natural language can be used to help guide IRL agents in unknown environments. 
While they showed that their method was effective, their work was limited in that the language that they investigated was highly structured as it was generated by a synthetic grammar. 
In addition, they limited their investigation to grid-based environments. 
In this work, we explore how natural language provided by humans can be used to improve learning in a complex environment that uses pixel information as state.

\section{Background}
\subsection{Reinforcement Learning}

Reinforcement learning (RL) is a machine learning technique where an agent's target is to solve a Markov Decision Process (MDP) by interacting with an environment through a trial and error process. 
A MDP can be expressed as a tuple of $<S,A,T,R, \gamma>$ where $S$ and $A$ are sets of possible states in an environment and actions an agent can take respectively. 
$T$ describes how actions transition the agent from one state to another. 
$R$ is a numeric reward function that describes the quality of a state.
$\gamma$ is known as the discount factor which determines how much emphasis an agent places on short-term versus long-term rewards. 
The goal of an agent in a MDP is to learn a policy $\pi$ that specifies the action that should be taken in each state that maximizes expected long-term reward.

\subsection{Proximal Policy Optimization}
Our A3PS algorithm builds on Proximal Policy Optimization (PPO)~\cite{ppo}. 
PPO formulates vanilla policy gradient in a way that provides more stable yet reliable action probabilities for a RL agent. 
Instead of using log probability to define the action's impact on the agent's current policy, importance of the action from current policy over the previous policy's action is measured (equation \ref{eq1}).
$r_t(\theta)$ differentiates between the old policy $\pi_{\theta_{old}}(a_t|s_t)$ and current policy $\pi_\theta(a_t|s_t)$.
Then clipping is done on the estimated advantage function $\hat{A}_t$ to avoid choosing the most expected actions for current policy (equation \ref{eq2}).                                                                                                                                                                                                                                                 
\begin{eqnarray}\label{eq1}
r_t(\theta)  =  \frac{\pi_\theta(a_t|s_t)}{\pi_{\theta_{old}}(a_t|s_t)}
\end{eqnarray}

\begin{eqnarray}\label{eq2}
L^{Clip}(\theta) = \hat{E}_t [min(i_t(\theta)\hat{A}_t, clip (r_t(\theta), 1-\epsilon, 1+\epsilon)\hat{A}_t)]
\end{eqnarray}

\begin{figure*}[t]
\centering
\includegraphics[scale=.6]{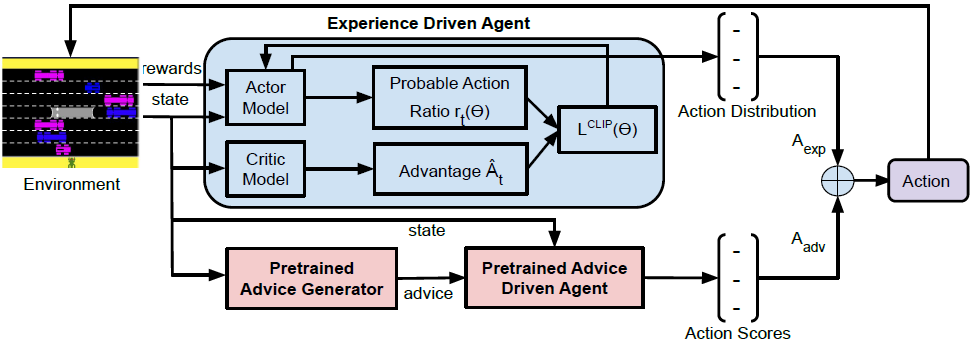}
\caption{Architecture of Automated Advice Aided Policy Shaping (A3PS)}
\label{fig_architecture}
\end{figure*}

\subsection{Policy Shaping}
The idea behind policy shaping is first introduced by Griffith \textit{et al.}~\cite{NIPS2013_5187,policy2} where both rewards and a numeric human feedback signal are combined to ultimately determine the agent's policy. 
Policy shaping does this by modeling the human's feedback policy as a probability distribution over potential actions. 
By modeling the human feedback policy in this way, it is possible to merge it with the agent's action policy. 
This mixed policy, thus, takes into account elements of both environmental reward and human feedback. 
In our work, we are employing a variant of policy shaping. 
Instead of human involvement, the human's feedback policy is determined through natural language advice by an agent rather than a numeric feedback signal. 
There are two primary advantages to this technique. 
The first of these is that language is a more natural way for humans to provide guidance.
The second is that language should help the agent generalize this advice over many states. 

\section{Methods}

\begin{figure}[t]
\centering
\includegraphics[width=\columnwidth]{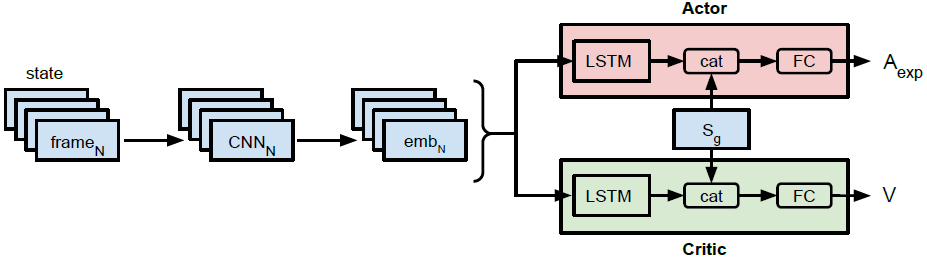}
\caption{Network architecture of Experience Driven RL Agent}
\label{fig_exp_agent}
\end{figure}

In this paper, we propose a variant policy shaping method, \textit{\textbf{A}utomated \textbf{A}dvice \textbf{A}ided \textbf{P}olicy \textbf{S}haping \textbf{(A3PS)}} (see Figure \ref{fig_architecture}). 
By combining human advice with RL, our method should better enable agents to learn in complex environments and environments with sparse rewards. 
Our proposed network A3PS is composed of three main modules: the Experience Driven Agent (EDA), the Advice Generator, and the Advice Driven Agent (ADA). 
The EDA is a Reinforcement learning agent that uses PPO to learn a distribution over future actions that maximizes expected future rewards. 
Thus, the EDA encapsulates knowledge learned through experience and is primarily driven by environmental reward. 
The ADA is a multi-modal deep neural network that uses pixel-based game state and advice texts as inputs from the pretrained advice generator and is responsible for producing the action score vector.
The action score vector is meant to represent the utility of actions based on human advice. 
The output of these modules are then combined to produce a final distribution that combines the knowledge gained from both human advice and the agent's own experience. 
This is then used to guide an agent during exploration in RL. 
We discuss each of these modules in greater detail below.

\subsection{Experience Driven RL Agent (EDA)}
The Experience Driven Agent (EDA) is a reinforcement learning agent implemented using the Proximal Policy Optimization (PPO) algorithm. 
The network architecture of the EDA is shown in Figure~\ref{fig_exp_agent}.
At every time step, it takes two inputs: 1) Four consecutive frames of the environment ($S_{t1},..S_{t4}$) and 2) A vector $S_g$ representing the status of completion of all the intermediate and final goals in the environment. 
We take 4 consecutive frames, so that the network can better capture the motion and direction of the environment objects. 
Each frame of input is passed through a Convolutional Neural Network that encodes the image frame to an embedding vector labeled as $emb$ in Figure~\ref{fig_exp_agent}.
The goal state vector $S_g$ keeps the agent updated about its completed goals and remaining goals. 
For each iteration, whenever a goal position is explored by the PPO agent, flag for the goal is set to 1. 
This information encourages the agent to remain task oriented in situations where subgoals may be repeatable.


The Actor-Critic model of EDA uses two deep neural networks: An Actor network gives action distribution from state and A Critic network tries to estimate the value for state-action combination. 
In our case, the state is composed of embedding vectors $emb_1..emb_4$. 
The four embedding vectors are sent to a LSTM~\cite{sepp1997lstm} iteratively in both Actor and Critic networks. 
The final output of the LSTM is concatenated with the goal state vector $S_g$ before sending it to a fully connected(FC) layer. 
The FC layer of Actor model populates action probabilities which assist the calculation of action ratio $r_t(\theta)$ (equation \ref{eq1}) and the Critic model generates a value which takes part in estimating the advantage function $\hat{A}_t$ of PPO algorithm.
Later PPO with clipped objective is applied to the estimated $\hat{A}_t$ to avoid drastically changing the policy.
Equation \ref{eq2} refers to the mathematical representation of the clipping process. 
In each iteration, this module generates action distribution vector $A_{exp}$ as outcome.





\subsection{Advice Generator}
One of the issues present in \cite{harrison2018guiding} is that it was difficult to determine which advice has to be applied to a state at any given time. 
In our method, we address this limitation by using the advice generator.
The Advice Generator module is a deep neural network which accepts the game state(current frame) as input and generates advice regarding the state. As the task is similar to image captioning, we adopt the image captioning model utilized in \cite{show} to implement the Advice generator. 
Our advice generator is trained using a paired dataset of environment states and human generated advice utterances. 

\subsection{Advice Driven Agent (ADA)}

\begin{figure}[t]
\centering
\includegraphics[width=\columnwidth]{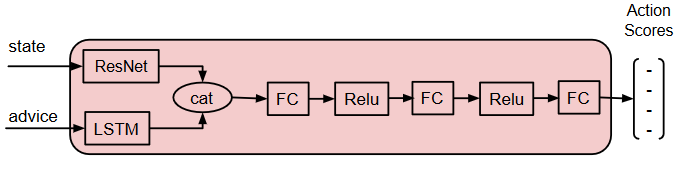}
\caption{Network architecture of Advice Driven Agent}
\label{fig_adv}
\end{figure}
The ADA agent utilizes the advice generated from the advice generator combined with state information to learn an action distribution based on human feedback. 
Figure \ref{fig_adv} shows how we design this agent.
In this work, we use Resnet-101 module to extracts the features from each state frame.
Pretrained GloVe embeddings~\cite{glove2014} are used to convert the text generated by the advice generator into vector representations. 
From these embedded words, LSTM cells capture advice context information in vector form.
Then these two vectors of Resnet and LSTM features are concatenated and advanced through multiple linear layers and ReLu activation layers to gather the decoded action scores $A_{adv}$ of the ADA module. 

\subsection{Automated Advice Aided Policy Shaping (A3PS)}

Figure \ref{fig_architecture} shows the overall architecture of A3PS.
As you can see in the figure, environmental state is used as an input to all three modules. 
These inputs are then processed via their respective networks to produce the action distributions $A_{exp}$ and $A_{adv}$ which we have already discussed in the previous segments of this section.
From the two predicted action distributions, the final action distribution is calculated using the equation \ref{eq7}. 
Here $\alpha$ is a weight variable to control how much weight will be given to the results of individual network. 
We decay the value of $\alpha$ over the training iterations. 
That means at starting point, the pretrained ADA network gets more importance with higher $\alpha$ value, but as the agent keeps exploring, $(1-\alpha)$ gets higher, hence EDA gains priority over the ADA module. 
%
Thus, the agent will prioritize human advice early during exploration and then rely on its own experience later on. 

\begin{eqnarray}\label{eq7}
a  =  softmax( \alpha * A_{adv}  + (1 - \alpha) *  A_{exp} )
\end{eqnarray}

At each time step, the agent chooses the most probable action according to this combined action distribution to execute. 
A note to mention, during training the A3PS architecture, pretrained networks for both the advice generator and ADA are utilized, so weights of the parameters of these two modules remain unchanged. 
Only the EDA is updated during RL as it is the only module that relies on environmental reward.

\section{Experimental Setup}

\subsection{Game Environment}

\begin{figure}[t]
\centering
\includegraphics[width=\columnwidth]{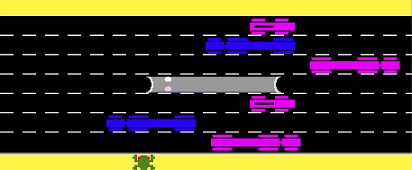}
\caption{Frogger game state with generated advice ``moved left get better position next move forward get around tunnel".}
\label{game}
\end{figure}

We test the A3PS system in the arcade game, Frogger (see Figure~\ref{game}). 
We chose Frogger because the game is easy enough for humans to provide high quality feedback while still being somewhat difficult for a RL agent to solve.
In Frogger, an agent must move from the bottom of the level to the top of the level while dodging car obstacles that approach from the left and right of the screen. 
In total, there are eight rows in the environment before the agent reaches the goal. 
The agent can move left, right, up, down, or choose to take no action at any time step. 
Also included in the environment is a tunnel in the center row that blocks the agent from moving through it. Cars, however, approaching on that row can move through it. 
For training purpose for all modules, the environmental state consists of the RGB values for 100x100 resized images of the game environment. 

\subsection{Reward Function}
We assign the highest reward of $+400$ if the agent reaches the goal in the game environment. 
The agent also receives rewards if it reaches certain rows for the first time. 
For example, once the agent reaches the second row in the environment for the first time, the agent will receive $+10$ reward. 
In addition, as there is an obstacle in level 5, a reward of $+100$ has given if the agent is able to reach the level by successfully overcoming the tunnel.
In addition, whenever an agent goes one level up regardless of whether the agent has performed same action before, $+1$ is rewarded.

Penalties are given to the agent for taking the wait action. 
If the agent waits in the starting row, $-5$ reward is given to the agent. 
The wait action results in $-1$ reward otherwise. 
If the agent moves off the side of the environment or the tunnel, then $-2$ is rewarded. 
If the agent is hit by a car, the agent receives $-20$ reward and the episode ends. 


\begin{figure}[t]
\centering
\includegraphics[width=\columnwidth]{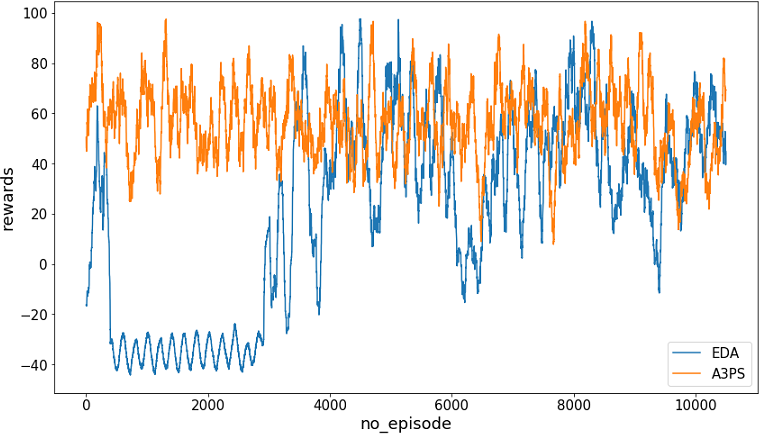}
\caption{Average episode reward for EDA and A3PS in dense reward setting (smoothed with 100 episodes moving window).}
\label{comparison}
\end{figure}

\begin{figure}[t]
\centering
\includegraphics[width=\columnwidth]{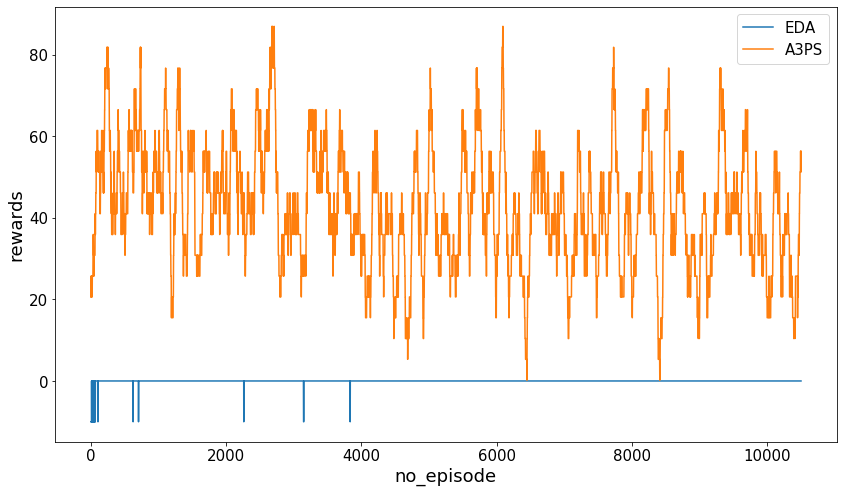}
\caption{Average episode reward for EDA and A3PS in sparse reward setting (smoothed with 100 episodes moving window).}
\label{comparison_sp_reward}
\end{figure}

\subsection{Dataset}

To train the A3PS agent, we require a corpus of human advice describing actions to take in various game states. 
In this paper, we utilize the dataset used in \cite{frog}. 
This dataset contains examples of state/action information paired with natural language explanations about why an action should be performed in a given state gathered from users on Mechanical Turk. 
In total, the dataset contains 1935 unique examples.
To train the ADA, we split this data into two parts using a ratio of 90 by 10. 
90 percent of the dataset contains 1741 examples and is used for training the advice related modules. 
The remaining 10 percent is used for parameter tuning these models. 
Each piece of human advice contained in this dataset is preprocessed before utilizing them in the training process. 
After discarding the special characters from the natural language texts, the NLTK-tokenizer is applied to each line of the texts. 
Then stop words are removed before adding the texts in the vocabulary dictionary.
This removal task is done to make sure that our advice driven agent only focuses on the necessary elements of the natural language advice. 

\subsection{Network Parameters}
Adam optimizer is used in all the networks as the optimization algorithm.
However, as learning rate, for the EDA, $1e^{-4}$  works best. 
Similarly, the ADA module and advice generator use learning rate of $1e^{-3}$ and $4e^{-4}$ respectively.  
For all networks, the LSTM size is fixed to 512. 
And the value for image embedding size for EDA network is set to 512.
All the experiments have been done using 2 Nvidia GTX 1080Ti GPUs.

\section{Evaluation \& Discussion}

In our experiments, we compare against the EDA network with no access to language as a baseline. 
Thus, we are evaluating if our A3PS algorithm can utilize human advice to the point where it can outperform a baseline RL agent. 
We compare in the Frogger environment described earlier under two conditions. 
The first of these uses a dense reward function while the other uses a much more sparse reward function. 
This will allow us to see if the presence of language can make up for deficiencies in the environmental reward function to enable an agent to learn. 


\subsection{Experiment 1: Dense Reward}

For this experiment, we have trained both standalone EDA network and A3PS network for 10,500 episodes. 
Both of the models utilize the reward function that has been discussed in the Experimental Setup section. 
Also, the weight decaying procedure takes place after each 2000 episodes from the starting episode until around 6000 episodes and decay happens by 0.2 each time.
As a result of this decay strategy, the A3PS agent solely uses a policy derived from environmental reward for the last 4500 episodes.  
Figure \ref{comparison} depicts the visualization of experiment 1.
As can be seen from the figure, the A3PS very quickly learns a decent policy and can quickly begin optimizing that policy. 
This shows that the A3PS agent is able to successfully interpret the advice from human trainers and synthesize it into useful policy information.
On the contrary, the baseline EDA model struggles to build a policy to reach the goal for the starting 3000 episodes. 
Though the baseline model starts to learn from its experience which is visible at around episode 4000, it still takes some time before it matches the performance of the A3PS agent.
This shows how powerful natural language as a source of human guidance can be. 
The agent in this situation was able to utilize this human advice to make up for its lack of experience. 
As time went on, the agent was able to refine this policy using its experience to learn an overall better policy.

\subsection{Experiment 2: Sparse Reward}
In this experiment, we investigate the learning performance between the agents in the case of sparse or ill-defined reward function. 
For this experiment, we define the reward function differently than the first experiment. The agent will get a positive reward (+400) only when it reaches the goal and a negative -20 reward when it dies. 
No other rewards are given or deducted from the agent. 
From Figure \ref{comparison_sp_reward} we see that the baseline agent EDA initially gets some negative rewards but later it gets neither positive nor negative rewards. 
This implies that the agent tries to explore further into the environment, but soon dies and receives the negative reward. 
This causes the agent to prefer to take no action as receiving no reward is preferable to the possibility of a negative reward. 
This illustrates why environments with sparse rewards can be difficult for RL agents. 

In contrast, the A3PS agent is able to supplement this sparse reward with a denser signal from the ADA module. 
This encourages the agent to explore the environment and discover positive environmental reward. 
After 6000 episodes, the A3PS agent does not take guidance from the ADA module and chooses actions only based on its experience. 
As we see from Figure~\ref{comparison_sp_reward}, still the agent can reach goal with its learned policy. 
Though the agent does not get any immediate rewards for its actions, automated guidance helps it to stay on the direction and shape its policy to the optimal convergence where it can take right policy after not being guided by the ADA module as well. 
In contrast, in baseline EDA network, without the guidance, the policy converges into local optima and consequently agent does not explore to reach the goal.

\section{Conclusion}

In this paper we present A3PS, an IRL algorithm that utilizes natural language advice combined with environmental reward to train agents in complex environments. 
By using pixel based game states and associative advice texts, we have shown its efficacy in the arcade game, Frogger.
We evaluated A3PS system against a baseline deep-RL method and showed that A3PS outperforms it for both dense and sparse reward functions, showing the effectiveness of our model and natural language as a source of instruction for IRL methods.

\section{Acknowledgements}

This material is based upon work supported by the National Science Foundation under Grant No. 1849231.

\bibliographystyle{flairs}
\bibliography{frog.bib}

\end{document}